\documentclass[runningheads]{llncs}
\usepackage[T1]{fontenc}
\usepackage{graphicx}
\usepackage{amsmath}
\usepackage{amssymb}
\usepackage{cite}
\usepackage{algorithm}
\usepackage{algpseudocode}
\usepackage{subcaption} 
\usepackage{caption}
\usepackage{hyperref}
\usepackage{array}
\usepackage{amsmath}
\usepackage{pgfplots}
\usepackage{subcaption}
\pgfplotsset{compat=1.18}
\usepackage{float}

\usepackage{pgfplots}
\usetikzlibrary{intersections}
\usepackage{bm}
\usetikzlibrary{positioning}
\usepgfplotslibrary{fillbetween}
\pgfplotsset{compat=1.5}
\usetikzlibrary{pgfplots.groupplots}
\usetikzlibrary{shapes.geometric,backgrounds}

\usepackage{tikz}
\usepackage{xifthen}

\pgfdeclarelayer{bg}    
\pgfsetlayers{bg,main}  
\definecolor{beige}{RGB}{245, 245, 220}

\definecolor{darkgrey}{RGB}{75, 75, 75}
\definecolor{lightgrey}{RGB}{250, 250, 250}

\usetikzlibrary{calc, arrows, fit, positioning, patterns, decorations.pathreplacing, shapes}
\tikzstyle{dash} = [dashed, -latex,>=latex]
\tikzstyle{line} = [draw, -latex,>=latex]
\tikzstyle{smallbox} = [draw, minimum size=5.0mm]
\tikzstyle{box} = [draw, minimum size=7.0mm]
\tikzstyle{bigbox} = [draw, minimum size=10.0mm]
\tikzstyle{rectangle} = [draw, minimum width=10.0mm, minimum height=20.0mm]
\tikzstyle{switch} = [trapezium, trapezium angle=120, draw, rotate=90,  inner ysep=5pt, outer sep=5pt,
minimum height=7mm, minimum width=7mm]
\tikzstyle{roundbox} = [draw, circle, inner sep=0pt, minimum size=3mm]
\tikzstyle{clamped} = [draw, fill=darkgrey, minimum size=0.15cm]
\tikzstyle{msgcircle} = [shape=circle, draw, inner sep=0pt, minimum size=4mm, fill=white, font=\scriptsize]
\tikzstyle{darkmsgcircle} = [shape=circle, draw, inner sep=0pt, minimum size=4mm, fill=darkgrey, text=white, font=\scriptsize]
\tikzstyle{redmsgcircle} = [shape=circle, draw=red, inner sep=0pt, minimum size=4mm, text=red, font=\scriptsize]
\tikzstyle{reddarkmsgcircle} = [shape=circle, draw=red, inner sep=0pt, minimum size=4mm, fill=red, text=white, font=\scriptsize]
\tikzstyle{msgdoublecircle} = [shape=circle, double, double distance=1.5pt, draw, inner sep=0pt, minimum size=5mm, fill=white]
\tikzstyle{darkmsgdoublecircle} = [shape=circle, double, double distance=1.5pt, draw, inner sep=0pt, minimum size=5mm, fill=darkgrey, text=white, font=\bfseries]

\newcommand{\msg}[6]{
      \ifthenelse{\isin{#1}{left} \AND \isin{#2}{down}}{
            \coordinate (anchor) at ($({#3})!{#5}!({#4})$);
            \node[xshift=-6.0mm] at (anchor) {#6};
            \node[xshift=-1.0mm] at (anchor) {$\downarrow$};
      }{}
      \ifthenelse{\isin{#1}{right} \AND \isin{#2}{down}}{
            \coordinate (anchor) at ($({#3})!{#5}!({#4})$);
            \node[xshift=6.0mm] at (anchor) {#6};
            \node[xshift=1.0mm] at (anchor) {$\downarrow$};
      }{}

      \ifthenelse{\isin{#1}{down} \AND \isin{#2}{right}}{
            \coordinate (anchor) at ($({#3})!{#5}!({#4})$);
            \node[ yshift=-4.0mm] at (anchor) {#6};
            \node[yshift=-1.0mm] at (anchor) {$\rightarrow$};
      }{}
      \ifthenelse{\isin{#1}{up} \AND \isin{#2}{right}}{
            \coordinate (anchor) at ($({#3})!{#5}!({#4})$);
            \node[ yshift=4.0mm] at (anchor) {#6};
            \node[yshift=1.0mm] at (anchor) {$\rightarrow$};
      }{}

      \ifthenelse{\isin{#1}{down} \AND \isin{#2}{left}}{
            \coordinate (anchor) at ($({#3})!{#5}!({#4})$);
            \node[ yshift=-4.0mm] at (anchor) {#6};
            \node[yshift=-1.0mm] at (anchor) {$\leftarrow$};
      }{}
      \ifthenelse{\isin{#1}{up} \AND \isin{#2}{left}}{
            \coordinate (anchor) at ($({#3})!{#5}!({#4})$);
            \node[ yshift=4.0mm] at (anchor) {#6};
            \node[yshift=1.0mm] at (anchor) {$\leftarrow$};
      }{}

      \ifthenelse{\isin{#1}{left} \AND \isin{#2}{up}}{
            \coordinate (anchor) at ($({#3})!{#5}!({#4})$);
            \node[ xshift=-6.0mm] at (anchor) {#6};
            \node[xshift=-1.0mm] at (anchor) {$\uparrow$};
      }{}
      \ifthenelse{\isin{#1}{right} \AND \isin{#2}{up}}{
            \coordinate (anchor) at ($({#3})!{#5}!({#4})$);
            \node[ xshift=6.0mm] at (anchor) {#6};
            \node[xshift=1.0mm] at (anchor) {$\uparrow$};
      }{}
}

\newcommand{\msgcircle}[6]{
      \ifthenelse{\isin{#1}{left} \AND \isin{#2}{down}}{
            \coordinate (anchor) at ($({#3})!{#5}!({#4})$);
            \node[msgcircle,xshift=-5.0mm] at (anchor) {#6};
            \node[xshift=-1.5mm] at (anchor) {$\downarrow$};
      }{}
      \ifthenelse{\isin{#1}{right} \AND \isin{#2}{down}}{
            \coordinate (anchor) at ($({#3})!{#5}!({#4})$);
            \node[msgcircle,xshift=5.0mm] at (anchor) {#6};
            \node[xshift=1.5mm] at (anchor) {$\downarrow$};
      }{}

      \ifthenelse{\isin{#1}{down} \AND \isin{#2}{right}}{
            \coordinate (anchor) at ($({#3})!{#5}!({#4})$);
            \node[msgcircle, yshift=-5.0mm] at (anchor) {#6};
            \node[yshift=-2.0mm] at (anchor) {$\rightarrow$};
      }{}
      \ifthenelse{\isin{#1}{up} \AND \isin{#2}{right}}{
            \coordinate (anchor) at ($({#3})!{#5}!({#4})$);
            \node[msgcircle, yshift=5.0mm] at (anchor) {#6};
            \node[yshift=2.0mm] at (anchor) {$\rightarrow$};
      }{}

      \ifthenelse{\isin{#1}{down} \AND \isin{#2}{left}}{
            \coordinate (anchor) at ($({#3})!{#5}!({#4})$);
            \node[msgcircle, yshift=-5.0mm] at (anchor) {#6};
            \node[yshift=-2.0mm] at (anchor) {$\leftarrow$};
      }{}
      \ifthenelse{\isin{#1}{up} \AND \isin{#2}{left}}{
            \coordinate (anchor) at ($({#3})!{#5}!({#4})$);
            \node[msgcircle, yshift=5.0mm] at (anchor) {#6};
            \node[yshift=2.0mm] at (anchor) {$\leftarrow$};
      }{}

      \ifthenelse{\isin{#1}{left} \AND \isin{#2}{up}}{
            \coordinate (anchor) at ($({#3})!{#5}!({#4})$);
            \node[msgcircle, xshift=-5.0mm] at (anchor) {#6};
            \node[xshift=-1.5mm] at (anchor) {$\uparrow$};
      }{}
      \ifthenelse{\isin{#1}{right} \AND \isin{#2}{up}}{
            \coordinate (anchor) at ($({#3})!{#5}!({#4})$);
            \node[msgcircle, xshift=5.0mm] at (anchor) {#6};
            \node[xshift=1.5mm] at (anchor) {$\uparrow$};
      }{}
}

\newcommand{\darkmsg}[6]{
      \ifthenelse{\isin{#1}{left} \AND \isin{#2}{down}}{
            \coordinate (anchor) at ($({#3})!{#5}!({#4})$);
            \node[darkmsgcircle, xshift=-5mm] at (anchor) {#6};
            \node[xshift=-1.5mm] at (anchor) {$\downarrow$};
      }{}
      \ifthenelse{\isin{#1}{right} \AND \isin{#2}{down}}{
            \coordinate (anchor) at ($({#3})!{#5}!({#4})$);
            \node[darkmsgcircle, xshift=5mm] at (anchor) {#6};
            \node[xshift=1.5mm] at (anchor) {$\downarrow$};
      }{}

      \ifthenelse{\isin{#1}{down} \AND \isin{#2}{right}}{
            \coordinate (anchor) at ($({#3})!{#5}!({#4})$);
            \node[darkmsgcircle, yshift=-5.0mm] at (anchor) {#6};
            \node[yshift=-2.0mm] at (anchor) {$\rightarrow$};
      }{}
      \ifthenelse{\isin{#1}{up} \AND \isin{#2}{right}}{
            \coordinate (anchor) at ($({#3})!{#5}!({#4})$);
            \node[darkmsgcircle, yshift=5.0mm] at (anchor) {#6};
            \node[yshift=2.0mm] at (anchor) {$\rightarrow$};
      }{}

      \ifthenelse{\isin{#1}{down} \AND \isin{#2}{left}}{
            \coordinate (anchor) at ($({#3})!{#5}!({#4})$);
            \node[darkmsgcircle, yshift=-5.0mm] at (anchor) {#6};
            \node[yshift=-2.0mm] at (anchor) {$\leftarrow$};
      }{}
      \ifthenelse{\isin{#1}{up} \AND \isin{#2}{left}}{
            \coordinate (anchor) at ($({#3})!{#5}!({#4})$);
            \node[darkmsgcircle, yshift=5.0mm] at (anchor) {#6};
            \node[yshift=2.0mm] at (anchor) {$\leftarrow$};
      }{}

      \ifthenelse{\isin{#1}{left} \AND \isin{#2}{up}}{
            \coordinate (anchor) at ($({#3})!{#5}!({#4})$);
            \node[darkmsgcircle, xshift=-5.0mm] at (anchor) {#6};
            \node[xshift=-1.5mm] at (anchor) {$\uparrow$};
      }{}
      \ifthenelse{\isin{#1}{right} \AND \isin{#2}{up}}{
            \coordinate (anchor) at ($({#3})!{#5}!({#4})$);
            \node[darkmsgcircle, xshift=5.0mm] at (anchor) {#6};
            \node[xshift=1.5mm] at (anchor) {$\uparrow$};
      }{}
}

\newcommand{\redbackmsg}[6]{
      \ifthenelse{\isin{#1}{left} \AND \isin{#2}{down}}{
            \coordinate (anchor) at ($({#3})!{#5}!({#4})$);
            \node[reddarkmsgcircle, xshift=-5mm] at (anchor) {#6};
            \node[xshift=-1.5mm] at (anchor) {$\downarrow$};
      }{}
      \ifthenelse{\isin{#1}{right} \AND \isin{#2}{down}}{
            \coordinate (anchor) at ($({#3})!{#5}!({#4})$);
            \node[reddarkmsgcircle, xshift=5mm] at (anchor) {#6};
            \node[xshift=1.5mm] at (anchor) {$\downarrow$};
      }{}

      \ifthenelse{\isin{#1}{down} \AND \isin{#2}{right}}{
            \coordinate (anchor) at ($({#3})!{#5}!({#4})$);
            \node[reddarkmsgcircle, yshift=-5.0mm] at (anchor) {#6};
            \node[yshift=-2.0mm] at (anchor) {$\rightarrow$};
      }{}
      \ifthenelse{\isin{#1}{up} \AND \isin{#2}{right}}{
            \coordinate (anchor) at ($({#3})!{#5}!({#4})$);
            \node[reddarkmsgcircle, yshift=5.0mm] at (anchor) {#6};
            \node[yshift=2.0mm] at (anchor) {$\rightarrow$};
      }{}

      \ifthenelse{\isin{#1}{down} \AND \isin{#2}{left}}{
            \coordinate (anchor) at ($({#3})!{#5}!({#4})$);
            \node[reddarkmsgcircle, yshift=-5.0mm] at (anchor) {#6};
            \node[yshift=-2.0mm] at (anchor) {$\leftarrow$};
      }{}
      \ifthenelse{\isin{#1}{up} \AND \isin{#2}{left}}{
            \coordinate (anchor) at ($({#3})!{#5}!({#4})$);
            \node[reddarkmsgcircle, yshift=5.0mm] at (anchor) {#6};
            \node[yshift=2.0mm] at (anchor) {$\leftarrow$};
      }{}

      \ifthenelse{\isin{#1}{left} \AND \isin{#2}{up}}{
            \coordinate (anchor) at ($({#3})!{#5}!({#4})$);
            \node[reddarkmsgcircle, xshift=-5.0mm] at (anchor) {#6};
            \node[xshift=-1.5mm] at (anchor) {$\uparrow$};
      }{}
      \ifthenelse{\isin{#1}{right} \AND \isin{#2}{up}}{
            \coordinate (anchor) at ($({#3})!{#5}!({#4})$);
            \node[reddarkmsgcircle, xshift=5.0mm] at (anchor) {#6};
            \node[xshift=1.5mm] at (anchor) {$\uparrow$};
      }{}
}

\newcommand{\redmsg}[6]{
      \ifthenelse{\isin{#1}{left} \AND \isin{#2}{down}}{
            \coordinate (anchor) at ($({#3})!{#5}!({#4})$);
            \node[redmsgcircle, xshift=-5mm] at (anchor) {#6};
            \node[xshift=-1.5mm] at (anchor) {$\downarrow$};
      }{}
      \ifthenelse{\isin{#1}{right} \AND \isin{#2}{down}}{
            \coordinate (anchor) at ($({#3})!{#5}!({#4})$);
            \node[redmsgcircle, xshift=5mm] at (anchor) {#6};
            \node[xshift=1.5mm] at (anchor) {$\downarrow$};
      }{}

      \ifthenelse{\isin{#1}{down} \AND \isin{#2}{right}}{
            \coordinate (anchor) at ($({#3})!{#5}!({#4})$);
            \node[redmsgcircle, yshift=-5.0mm] at (anchor) {#6};
            \node[yshift=-2.0mm] at (anchor) {$\rightarrow$};
      }{}
      \ifthenelse{\isin{#1}{up} \AND \isin{#2}{right}}{
            \coordinate (anchor) at ($({#3})!{#5}!({#4})$);
            \node[redmsgcircle, yshift=5.0mm] at (anchor) {#6};
            \node[yshift=2.0mm] at (anchor) {$\rightarrow$};
      }{}

      \ifthenelse{\isin{#1}{down} \AND \isin{#2}{left}}{
            \coordinate (anchor) at ($({#3})!{#5}!({#4})$);
            \node[redmsgcircle, yshift=-5.0mm] at (anchor) {#6};
            \node[yshift=-2.0mm] at (anchor) {$\leftarrow$};
      }{}
      \ifthenelse{\isin{#1}{up} \AND \isin{#2}{left}}{
            \coordinate (anchor) at ($({#3})!{#5}!({#4})$);
            \node[redmsgcircle, yshift=5.0mm] at (anchor) {#6};
            \node[yshift=2.0mm] at (anchor) {$\leftarrow$};
      }{}

      \ifthenelse{\isin{#1}{left} \AND \isin{#2}{up}}{
            \coordinate (anchor) at ($({#3})!{#5}!({#4})$);
            \node[redmsgcircle, xshift=-5.0mm] at (anchor) {#6};
            \node[xshift=-1.5mm] at (anchor) {$\uparrow$};
      }{}
      \ifthenelse{\isin{#1}{right} \AND \isin{#2}{up}}{
            \coordinate (anchor) at ($({#3})!{#5}!({#4})$);
            \node[redmsgcircle, xshift=5.0mm] at (anchor) {#6};
            \node[xshift=1.5mm] at (anchor) {$\uparrow$};
      }{}
}

\newcommand{\bwmsg}[6]{
      \ifthenelse{\isin{#1}{left} \AND \isin{#2}{down}}{
            \coordinate (anchor) at ($({#3})!{#5}!({#4})$);
            \node[msgdoublecircle, xshift=-5.5mm] at (anchor) {#6};
            \node[xshift=-1.5mm] at (anchor) {$\downarrow$};
      }{}
      \ifthenelse{\isin{#1}{right} \AND \isin{#2}{down}}{
            \coordinate (anchor) at ($({#3})!{#5}!({#4})$);
            \node[msgdoublecircle, xshift=5.5mm] at (anchor) {#6};
            \node[xshift=1.5mm] at (anchor) {$\downarrow$};
      }{}

      \ifthenelse{\isin{#1}{down} \AND \isin{#2}{right}}{
            \coordinate (anchor) at ($({#3})!{#5}!({#4})$);
            \node[msgdoublecircle, yshift=-6.0mm] at (anchor) {#6};
            \node[yshift=-2.0mm] at (anchor) {$\rightarrow$};
      }{}
      \ifthenelse{\isin{#1}{up} \AND \isin{#2}{right}}{
            \coordinate (anchor) at ($({#3})!{#5}!({#4})$);
            \node[msgdoublecircle, yshift=6.0mm] at (anchor) {#6};
            \node[yshift=2.0mm] at (anchor) {$\rightarrow$};
      }{}

      \ifthenelse{\isin{#1}{down} \AND \isin{#2}{left}}{
            \coordinate (anchor) at ($({#3})!{#5}!({#4})$);
            \node[msgdoublecircle, yshift=-6.0mm] at (anchor) {#6};
            \node[yshift=-2.0mm] at (anchor) {$\leftarrow$};
      }{}
      \ifthenelse{\isin{#1}{up} \AND \isin{#2}{left}}{
            \coordinate (anchor) at ($({#3})!{#5}!({#4})$);
            \node[msgdoublecircle, yshift=6.0mm] at (anchor) {#6};
            \node[yshift=2.0mm] at (anchor) {$\leftarrow$};
      }{}

      \ifthenelse{\isin{#1}{left} \AND \isin{#2}{up}}{
            \coordinate (anchor) at ($({#3})!{#5}!({#4})$);
            \node[msgdoublecircle, xshift=-5.5mm] at (anchor) {#6};
            \node[xshift=-1.5mm] at (anchor) {$\uparrow$};
      }{}
      \ifthenelse{\isin{#1}{right} \AND \isin{#2}{up}}{
            \coordinate (anchor) at ($({#3})!{#5}!({#4})$);
            \node[msgdoublecircle, xshift=5.5mm] at (anchor) {#6};
            \node[xshift=1.5mm] at (anchor) {$\uparrow$};
      }{}
}

\newcommand{\bwdarkmsg}[6]{
      \ifthenelse{\isin{#1}{left} \AND \isin{#2}{down}}{
            \coordinate (anchor) at ($({#3})!{#5}!({#4})$);
            \node[darkmsgdoublecircle, xshift=-5.5mm] at (anchor) {#6};
            \node[xshift=-1.5mm] at (anchor) {$\downarrow$};
      }{}
      \ifthenelse{\isin{#1}{right} \AND \isin{#2}{down}}{
            \coordinate (anchor) at ($({#3})!{#5}!({#4})$);
            \node[darkmsgdoublecircle, xshift=5.5mm] at (anchor) {#6};
            \node[xshift=1.5mm] at (anchor) {$\downarrow$};
      }{}

      \ifthenelse{\isin{#1}{down} \AND \isin{#2}{right}}{
            \coordinate (anchor) at ($({#3})!{#5}!({#4})$);
            \node[darkmsgdoublecircle, yshift=-6.0mm] at (anchor) {#6};
            \node[yshift=-2.0mm] at (anchor) {$\rightarrow$};
      }{}
      \ifthenelse{\isin{#1}{up} \AND \isin{#2}{right}}{
            \coordinate (anchor) at ($({#3})!{#5}!({#4})$);
            \node[darkmsgdoublecircle, yshift=6.0mm] at (anchor) {#6};
            \node[yshift=2.0mm] at (anchor) {$\rightarrow$};
      }{}

      \ifthenelse{\isin{#1}{down} \AND \isin{#2}{left}}{
            \coordinate (anchor) at ($({#3})!{#5}!({#4})$);
            \node[darkmsgdoublecircle, yshift=-6.0mm] at (anchor) {#6};
            \node[yshift=-2.0mm] at (anchor) {$\leftarrow$};
      }{}
      \ifthenelse{\isin{#1}{up} \AND \isin{#2}{left}}{
            \coordinate (anchor) at ($({#3})!{#5}!({#4})$);
            \node[darkmsgdoublecircle, yshift=6.0mm] at (anchor) {#6};
            \node[yshift=2.0mm] at (anchor) {$\leftarrow$};
      }{}

      \ifthenelse{\isin{#1}{left} \AND \isin{#2}{up}}{
            \coordinate (anchor) at ($({#3})!{#5}!({#4})$);
            \node[darkmsgdoublecircle, xshift=-5.5mm] at (anchor) {#6};
            \node[xshift=-1.5mm] at (anchor) {$\uparrow$};
      }{}
      \ifthenelse{\isin{#1}{right} \AND \isin{#2}{up}}{
            \coordinate (anchor) at ($({#3})!{#5}!({#4})$);
            \node[darkmsgdoublecircle, xshift=5.5mm] at (anchor) {#6};
            \node[xshift=1.5mm] at (anchor) {$\uparrow$};
      }{}
}

\tikzset{mainstyle/.style={fill=white, draw=black, shape=rectangle, align=center}}

\tikzset{dstyle/.style={mainstyle, minimum size=4mm, inner sep=0pt, text width=4mm}}

\tikzset{sstyle/.style={mainstyle, minimum size=5mm, inner sep=0pt, text width=5mm}}

\tikzset{ostyle/.style={fill=darkgrey, draw=black, shape=rectangle, minimum size=0.2cm, inner sep=0pt, text width=2mm}}

\tikzstyle{observation}=[ostyle]
\tikzstyle{deterministic}=[dstyle]
\tikzstyle{stochastic}=[sstyle]

\tikzstyle{filter}=[mainstyle, minimum width=1cm, minimum height=0.5cm]
\tikzstyle{selector}=[fill=white, draw=black, shape=trapezium, rotate=180, minimum width=1cm, minimum height=0.5cm]

\DeclareRobustCommand{\cev}[1]{%
  \mathpalette\do@cev{#1}%
}
\newcommand{\do@cev}[2]{%
  \fix@cev{#1}{+}%
  \reflectbox{$\m@th#1\vec{\reflectbox{$\fix@cev{#1}{-}\m@th#1#2\fix@cev{#1}{+}$}}$}%
  \fix@cev{#1}{-}%
}
\newcommand{\fix@cev}[2]{%
  \ifx#1\displaystyle
    \mkern#23mu
  \else
    \ifx#1\textstyle
      \mkern#23mu
    \else
      \ifx#1\scriptstyle
        \mkern#22mu
      \else
        \mkern#22mu
      \fi
    \fi
  \fi
}

\usepackage[disable]{todonotes}

\begin{document}
\title{Spike-based Belief Propagation in Nonlinear Dynamical Systems}
%
%

\author{Sepideh Adamiat\inst{1}\orcidID{0009-0009-7269-1596}
\and Hongye Wang\inst{2}
\and Wouter M. Kouw\inst{1}\orcidID{0000-0002-0547-4817}
\and Bert de Vries\inst{1}\orcidID{0000-0003-0839-174X}}
\authorrunning{S. Adamiat et al.}

\institute{
Electrical Engineering Department, Eindhoven University of Technology, Eindhoven, the Netherlands
\and
Biomedical Signals and Systems, Faculty of Electrical Engineering, Mathematics and Computer Science (EEMCS), University of Twente, Enschede, The Netherlands
}

\maketitle              
\begin{abstract}
\begin{sloppypar}
This paper presents a Bayesian control framework that integrates spike-based dynamics with probabilistic inference for adaptive control. Bayesian inference is widely regarded as a core computational principle of brain function, providing a normative framework for perception, decision-making, and learning under uncertainty. By combining a biologically inspired spiking neural model with Bayesian inference principles, we propose a brain-like control algorithm capable of operating in uncertain environments. We use the mountain car parking problem as a benchmark with non-linear dynamics. Our results demonstrate that the proposed controller can successfully update states in real time and generate goal-directed action plans through spike-driven dynamics. The results highlight the proposed model's potential as a bridge between computational neuroscience and probabilistic control theory.
\end{sloppypar}


\end{abstract}

\section{Introduction}\label{sec:intro}

Modeling the human brain remains a central challenge in modern science, motivated in part by the remarkable efficiency of biological intelligence. Humans learn and adapt using far less energy and data than contemporary artificial systems; for instance, the brain operates on roughly 20 Watts \cite{yu2018evaluating}, whereas modern AI systems typically require many orders of magnitude more computational resources.

Computational neuroscience has evolved from detailed biophysical simulations to theoretical frameworks explaining higher-level cognition. These approaches are broadly categorized as top-down and bottom-up. Top-down methods, such as the Bayesian brain hypothesis \cite{doya2007bayesian}, and the free energy principle \cite{friston2006free}, 
provide normative accounts of perception and decision-making, where normative refers to specifying how an ideal agent should optimally infer and act under uncertainty. However, such approaches often abstract away biological constraints. 
Bottom-up approaches, including spiking neural networks (SNNs) \cite{gerstner2002spiking} emphasize neural dynamics and biophysical realism. SNNs hold promise for improved energy efficiency due to their event-driven computation, where processing occurs only when spikes are generated, enabling sparse communication and reduced power consumption, particularly on neuromorphic hardware~\cite{roy2019towards}. 

Recent work aims to bridge these perspectives by developing brain-inspired algorithms that combine principled inference with neural implementation ~\cite{vafaii2024brain, steimer2009belief, bouttier2021circular, deneve2004bayesian}. In this paper, we follow this direction by integrating Bayesian inference with spiking neural networks.

 To enable scalable inference, we adopt a distributed message-passing formulation on factor graphs \cite{kschischang2002factor}, also known as Belief Propagation (BP). This approach decomposes global inference into local computations, aligning with the parallel and decentralized organization of neural systems. BP has been applied to nonlinear control and multi-agent systems \cite{van2019simulating, adamiat2024message, van2024multi} and is well-suited to event-driven (reactive) implementations \cite{bagaev2023reactive}.

Several studies have explored BP in SNNs \cite{bouttier2021circular}. A close relationship between BP and neurodynamic models was established in \cite{ott2006neurodynamics}. However, that work was limited to Markov random fields, which constitute a subset of factor graphs. In addition, the considered formulation focused on binary-valued variables. Steimer et al.~\cite{steimer2009belief} extended this approach by implementing BP in SNNs using a network of interconnected liquid state machines ~\cite{maass2011liquid}. Nevertheless, these approaches primarily focused on binary-valued variables.

Many real-world inference and control problems involve continuous-valued states, such as positions and velocities, requiring Gaussian belief representations. A spike-based framework for Gaussian BP was introduced in \cite{Adamiat2026spiking}, demonstrating belief updating through linear factor nodes and implementing two fundamental Bayesian inference problems: Kalman filtering and linear regression. However, the framework was limited to linear operations and did not support multivariate belief updates.

The main contributions of this paper are:
\begin{enumerate}
\item Integration of a spike-based node function into an event-driven BP framework, capable of performing state estimation and action planning in real time.
\item A spike-based message passing implementation of multivariate and nonlinear belief updates for the Mountain Car problem~\cite{sutton1998reinforcement}. 

\end{enumerate}

The proposed framework successfully solves the Mountain Car problem via message passing-based inference on a factor graph and provides a concrete step toward bridging the algorithmic level of probabilistic inference on factor graphs and their implementation via spiking neural dynamics.

\section{Problem Statement}\label{sec:problem}
\begin{figure}[t]
    \centering
    \includegraphics[width=\linewidth]{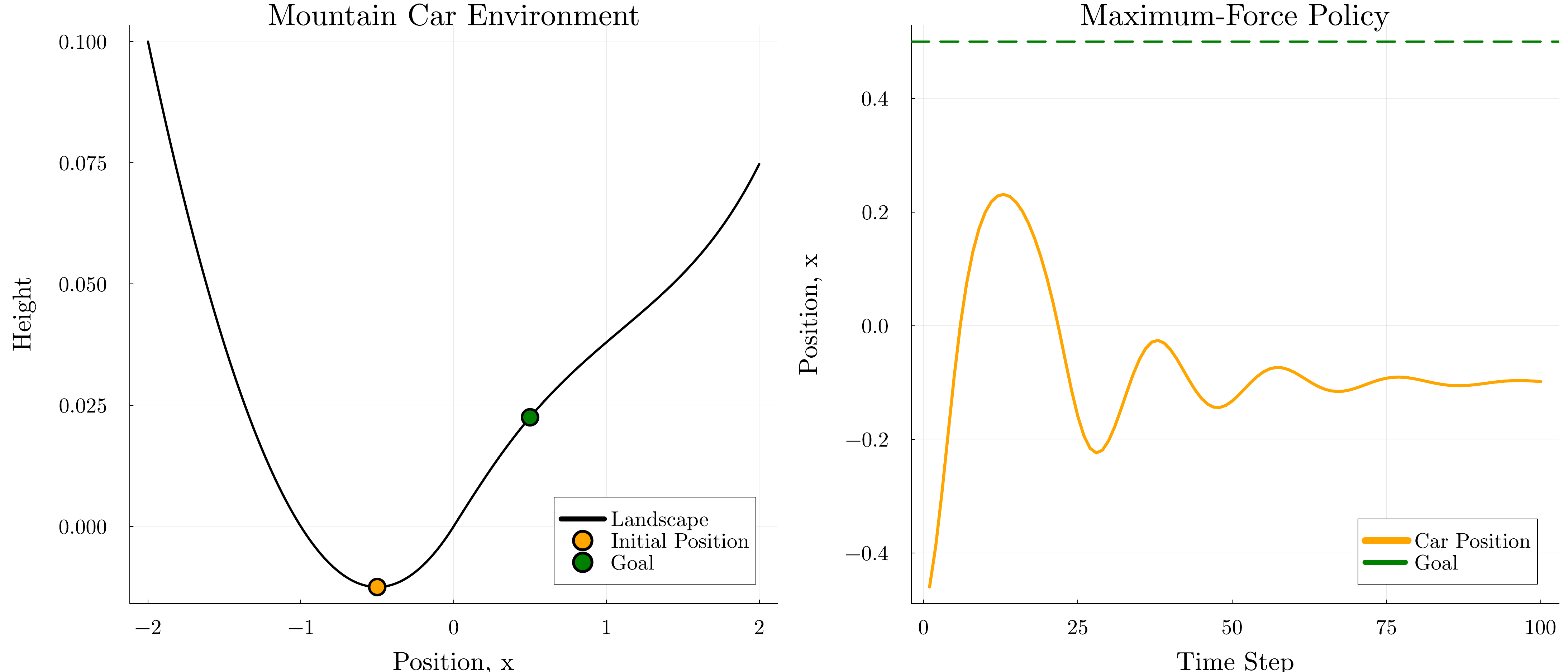}
    \caption{(left) The Mountain Car environment; (right) trajectory of the car position when the agent naively applies the maximum forward engine force at every time step. The orange curve shows the resulting position over time, while the dashed green line indicates the goal position. This strategy fails to reach the goal state; see Section~\ref{sec:problem}. }
    \label{fig:mountaincar}
\end{figure}

The Mountain Car problem is a classical benchmark in control theory and reinforcement learning. This environment consists of a car located in a valley between two hills, intending to reach a goal position. A schematic representation of the environment is shown in Fig.~\ref{fig:mountaincar} (left), illustrating the mountain profile and the car's position. In this paper, we use the formulation introduced in~\cite{ueltzhoffer2018deep}. 

In this problem, the vehicle is assumed to have a limited engine force. The car's trajectory when maximum force is continuously applied toward the goal is illustrated in Fig.~\ref{fig:mountaincar} (right). As observed, the car cannot reach the goal position due to insufficient power. 

To successfully reach the goal, the control policy must first drive the car in the opposite direction (toward the left hill) to build up sufficient momentum. The accumulated kinetic energy can then propel the car toward the right hill, ultimately reaching the goal position.

To solve this problem, we follow the probabilistic framework proposed in \cite{van2019simulating}. We consider an agent with an internal state defined as $s_t = [x_t, \dot{x_t}]$, where $x_t$ and $\dot{x_t}$ denote the position and velocity of the agent at time step $t$, respectively. At each time step, the agent applies an action $u_t$ to the environment and receives an observation $y_t$. The agent maintains probabilistic beliefs over its internal states, observations, and actions, all of which are modeled as Gaussian random variables.

To describe the relationship among hidden states, observations, and actions, we employ a generative model that provides a probabilistic description of how observations are generated. In this work, the probabilistic dependencies among the variables in the generative model are represented graphically using factor graphs. A biological agent continuously exchanges information with its environment and updates the internal states of its model to minimize expected surprise about sensory observations~\cite{friston2013life}. 

Following this framework, we design a synthetic agent that continuously updates its beliefs about hidden states and control variables based on incoming observations, thereby selecting actions that increase the likelihood of achieving desired observations. These belief updates are performed using BP \cite{loeliger_factor_2007} on the factor graph representation of the generative model.

The primary objective of this work is to bridge the gap between BP on factor graphs and the spike-based dynamics observed in biological neural systems. The distributed and local computations performed by BP algorithms have motivated their study as plausible models of neural information processing and probabilistic inference~\cite{loeliger_factor_2007,friston2017graphical}. In particular, our main contribution is a spike-based implementation of the nonlinear and multivariate state-transition factor, $p(s_t \mid s_{t-1},u_t)$, that operates on Gaussian beliefs. This factor reflects the Mountain Car environment's physical dynamics and is one of the most challenging components of the BP procedure.

Accordingly, the main problems addressed in this work are: 
\begin{enumerate}
    \item a spike-based representation of belief parameters (Section~\ref{sub:encoding}).
    \item a spiking neural implementation of a nonlinear and multivariate state-transition factor (Section~\ref{subsection:SNN_node}).
    \item the integration of this neural factor into a BP-based inference framework  (Section~\ref{subsection:model-inference}).
\end{enumerate}

\section{Technical Background}\label{sec:background} 


This section presents the technical background required for the proposed model. Subsections~\ref{subsec:Factor},~\ref{subsec:inference},~\ref{subsec:control} describe the probabilistic representation, the inference algorithm, and the control algorithm, respectively. Subsections~\ref{sub:LIF},~\ref{sub:encoding},~\ref{sub:nef-networks} then introduce the neural dynamics, the encoding and decoding of beliefs in the spike domain, and the methods used to implement the desired functions using networks of spiking neurons.

\subsection{Model Representation by Factor Graphs}\label{subsec:Factor}

To enable efficient and scalable Bayesian inference, we adopt a factor graph representation, which expresses a factorized probabilistic model and supports inference via local message (belief) passing. This structure promotes modularity and aligns naturally with neural computation, where local interactions correspond to message exchanges.

We use Forney-style Factor Graphs (FFGs), in which edges represent variables and nodes represent factors~\cite{forney2001codes}. Since each edge connects at most two nodes, branching (equality) nodes are introduced to model variables that appear in more than two factors~\cite{loeliger2004introduction}, thus enabling a graphical representation for any factorized probabilistic model.  

\begin{figure}[bth]
\centering
\begin{tikzpicture}
    \node[box] (g1) {$g$};
    \node[smallbox, right = 8 mm of g1] (eq1) {$=$};
    \node[box, below = 6 mm of eq1] (dot1) {$h$};
    \node[clamped, above  = 7 mm of g1] (u1) {};
    \node[above = 0 mm of u1](u1_hat){$\delta(u_t-\hat{u}_t)$};
    \node[clamped, below = 8 mm of dot1] (y1) {};
    \node[below = 0 mm of y1](y1_hat){$\delta(y_t-\hat{y}_t)$};
    \node[box, left = 8 mm of g1] (prior) {$\mathcal{N}$};
    \node[above = 0 mm of prior] () {$p\left(s_{t-1}\right)$};
    \node[right = 8 mm of eq1] (g2) {};
    
    \draw[->] (g1) -- (eq1)node[pos=0.5, above] {$s_t'$};
    \draw[->] (u1) -- (g1)node[pos=0.5, right] {$u_t$};
    \draw[->] (prior) -- (g1)node[pos=0.5, above] {};
    \draw[->] (eq1) -- (dot1)node[pos=0.5, right] {$s_t''$};
    \draw[->] (dot1) -- (y1)node[pos=0.5, right] {$y_t$};

    \draw[->] (eq1) -- (g2)node[pos=0.5, above] {$s_t$};

\end{tikzpicture}
\caption{An FFG representation of one time step of the state space model defined in \eqref{eq:state_update}, constrained by $u_t = \hat{u}_t$ and $y_t = \hat{y}_t$.}
\label{fig:FFG-state_update}
\end{figure}
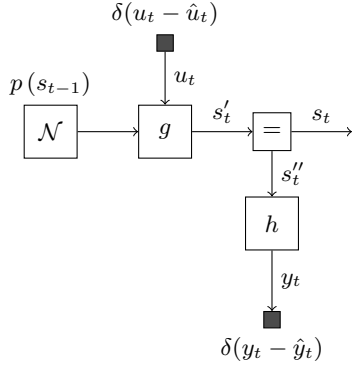

Consider a state space model,
\begin{equation} \label{eq:state_update}
p(s_t, y_t, u_t, s_{t-1}) = \underbrace{p(y_t \mid s_t)}_{\text{observation}}\, \underbrace{p(s_t \mid s_{t-1}, u_t)}_{\text{state transition}} \underbrace{p(u_t) p(s_{t-1})}_{ \text{priors}} \, , 
\end{equation}
which describes the relationship between successive states $s_{t-1}$ and $s_t$, outputs $y_t$ and control variables (actions) $u_t$. At time step $t$, after selecting an action $u_t = \hat{u}_t$ and having observed $y_t = \hat{y}_t$, the FFG representation of this model is shown in Fig.~\ref{fig:FFG-state_update}.

This FFG comprises four types of nodes. A data node (small black box) represents a delta constraint on a variable with a known value. For example, an observed variable $y_t$ with measured value $\hat{y}_t$ is represented by the factor $\delta(y_t-\hat{y}_t)$. A Gaussian factor node encodes a Gaussian distribution parameterized by its mean and covariance. An equality (branch) node enforces consistency among connected variables and enables Bayesian belief updating. The equality node for $s_t$ is given by
\begin{equation}
f_=(s_t,s_t',s_t'') = \delta(s_t-s_t')\delta(s_t-s_t'') .
\end{equation}
This equality node ensures that the posterior beliefs over $s_t$, $s_t'$, and $s_t''$ are identical.

Finally, the nodes $g$ and $h$ represent the state transition and observation models $p(s_t \mid s_{t-1}, u_t)$ and $p(y_t|s_t)$, respectively. 

Although FFG edges are inherently undirected, we adopt a fixed direction convention, which we later use to indicate the direction of message propagation in the graph.


\subsection{Inference by Message Passing}\label{subsec:inference}
We use a message passing algorithm to perform inference on an FFG. Messages propagate along edges and are updated locally in connected nodes. In this work, we employ the sum–product update rule~\cite{loeliger_factor_2007}. Since messages can propagate in both directions along an edge, we refer to messages traveling in the assigned edge direction as \emph{forward messages}, denoted by $\overrightarrow{\mu}$, and messages traveling in the opposite direction as \emph{backward messages}, denoted by $\overleftarrow{\mu}$.

In sum-product message passing, for a general factor node $f(y, x_1, \ldots, x_n)$, with incoming messages $\overrightarrow{\mu}_{x_i}(x_i)$, the outgoing message toward $y$ is given by \cite{kschischang2002factor,loeliger2004introduction}
\begin{equation} \underbrace{\overrightarrow\mu_y({y})}_{\text{outgoing message}}=\int\underbrace{\overrightarrow\mu_{x_1}({x}_1)...\overrightarrow\mu_{x_n}({x}_n) }_{\text{incoming messages}} \underbrace{f({y},{x}_1,...,{x}_n)}_{\text{node function}} \mathrm{d}{x}_1...\mathrm{d}{x}_n \, . \label{eq:sum-product} 
\end{equation}
This rule yields exact Bayesian inference results when the underlying graph is a tree.  

When two messages collide at an edge, the posterior marginal belief is obtained by multiplying the messages. More precisely, if the colliding messages are $\overrightarrow{\mu}_y(y)$ and $\overleftarrow{\mu}_y(y)$, the marginal belief over $y$ is given by
\begin{equation}
p(y) = \overrightarrow{\mu}_y(y)\,\overleftarrow{\mu}_y(y).
\end{equation}

For a more detailed explanation of FFGs and BP-based inference, we recommend~\cite{loeliger_factor_2007}.

%

\subsection{Control as Inference}\label{subsec:control}
Control-as-inference reformulates optimal control as a probabilistic inference problem, in which actions are obtained by inferring trajectories that are consistent with both the system dynamics and the desired outcomes \cite{kappen2012optimal}. This perspective enables the use of probabilistic graphical models and message-passing algorithms for planning and control.

\begin{figure}[bth]
\centering
\begin{tikzpicture}
    \node[box] (g1) {$g$};
    \node[smallbox, right = 8 mm of g1] (eq1) {$=$};
    \node[box, below = 6 mm of eq1] (dot1) {$h$};
    \node[box, above  = 7 mm of g1] (u1) {$\mathcal{N}$};
    \node[box, below = 8 mm of dot1] (y1) {$\mathcal{N}$};
    \node[below = 0 mm of y1] () {$p'(y_{t})$};
    \node[box, left = 8 mm of g1] (prior) {$\mathcal{N}$};
    \node[above = 0mm of prior] () {$p\left(s_{t-1}\right)$};
    
    \draw[->] (g1) -- (eq1);
    \draw[->] (u1) -- (g1)node[pos=0.5, right] {$u_t$};
    \draw[->] (prior) -- (g1)node[pos=0.5, above] {};
    \draw[->] (eq1) -- (dot1);
    \draw[->] (dot1) -- (y1)node[pos=0.5, right] {};

    \node[box, right = 10 mm of eq1] (g2) {$g$};
    \node[box, above  = 8 mm of g2] (BOX1) {$\mathcal{N}$};
    
    \node[smallbox, right = 12 mm of g2] (eq2) {$=$};
    \node[box, below = 6 mm of eq2] (dot2) {$h$};

    \node[box, below = 8 mm of dot2] (y2) {$\mathcal{N}$};
    \node[below = 0 mm of y2] () {$p'(y_{t+1})$};
    \node[right = 5 mm of eq2] (dots) {$...$};

    \draw[->] (eq1) -- (g2)node[pos=0.5, above] {$s_t$};
    \draw[->] (g2) -- (eq2)node[pos=0.5, above] {};;
    \draw[->] (BOX1) -- (g2)node[pos=0.5, right] {$u_{t+1}$};
    \draw[->] (eq2) -- (dot2);
    \draw[->] (dot2) -- (y2);
    \draw[->] (eq2) -- (dots)node[pos=0.9, above] {$s_{t+1}$};

    \node[box, right = 4 mm of dots] (gT) {$g$};
    \node[box, above  = 8 mm of gT] (BOXT) {$\mathcal{N}$};

    \node[box, below right = 6 mm and 10 mm of gT] (dotT) {$h$};
    \node[box, below  = 8 mm of dotT] (yT) {$\mathcal{N}$};
    \node[below = 0 mm of yT] () {$p'(y_{t+T})$};
    \node[left = 3 mm of gT] (leftT) {};
    
    \draw[->] (BOXT) -- (gT)node[pos=0.5, right] {$u_{t+T}$};

    \draw[->] (leftT) -- (gT);
    \draw[->] (gT) -| (dotT)node[pos=0.3, above] {$s_{t+T}$};
    \draw[->] (dotT) -- (yT);

\end{tikzpicture}
\caption{FFG of the extended model in~\eqref{eq:complete-model}. This model extends the state update model over a horizon of $T$ future time steps from time step $t$. The "control as inference" objective is to infer the posterior belief for $u_{t:t+T}$ conditioned on preferences $p'(y_{t:t+T})$.}
\label{fig:FFG-KF}
\end{figure}
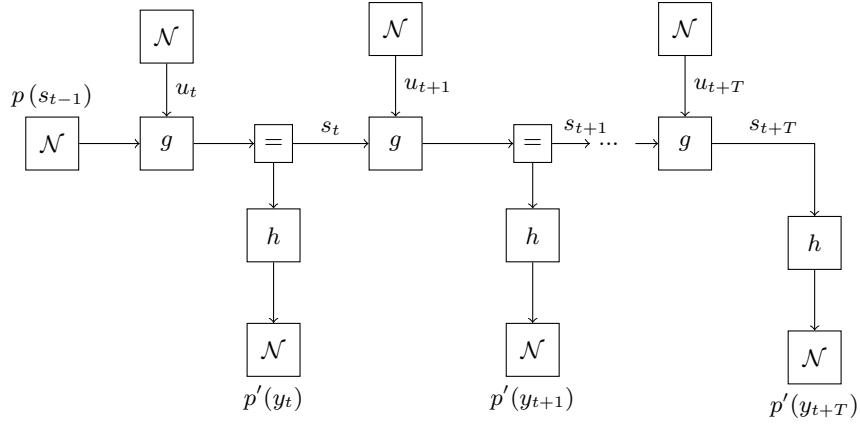

Active inference, a corollary of the Free Energy Principle \cite{friston2006free}, is a form of control-as-inference that originated in computational neuroscience as a theory of brain function. In this paper, we use the active inference model proposed in \cite{van2019simulating}. At each time step $t$, this model plans $T$ time steps into the future. In the context of the Mountain Car problem, this enables the agent to learn that it must explore the environment and select an appropriate sequence of actions to reach the goal. The model extends the state space model defined in~\eqref{eq:state_update} to future time steps by introducing control priors $p(u_k)$ and \emph{preferred} future observations, encoded by $p'(y_k)$. The corresponding extended generative model at the start of time step $t$ can be written as
\begin{equation}\label{eq:complete-model}
    p(y, s, u)
    \propto
    \underbrace{p(s_{t-1})}_{\substack{\text{state}\\ \text{prior}}}
    \prod_{k=t}^{t+T}
    \underbrace{p(y_k \mid s_k)}_{\text{observations}}\,
    \underbrace{p(s_k \mid s_{k-1}, u_k)}_{\text{state transition}}\,
    \underbrace{p(u_{k})}_{\substack{\text{control} \\ \text{prior}}}
    \underbrace{p'(y_{k})}_{\substack{\text{preferred} \\ \text{observations}}}\,,
\end{equation}
where $p(u_k)$ represents prior beliefs over a control (action) space, and $p'(y_k)$ encodes prior preferences ("goals") for future observations  \cite{adamiat2024message}. The node $p(s_{t-1})$  represents updated beliefs $p(s_{t-1}|y_{1:t-1})$ after having observed $y_{1:t-1}$. The FFG of this model is represented in Figure~\ref{fig:FFG-KF}.

The active inference approach to control does not require an explicit reward function. Instead, rewarding future states (or future outputs, since unobserved future outputs are also latent states) is encoded as an extension of the generative model; consequently, regular inference on this extended model will yield actions that fulfill the preferred future states.

\subsection{The Leaky Integrate-and-Fire Neuron Model}\label{sub:LIF}

This study used the Leaky Integrate-and-Fire (LIF) model\cite{gerstner2002spiking}. The LIF model simplifies the complex behavior of biological neurons while
retaining the essential aspects. The model is based on the membrane potential $V(t)$ at time $t$, whose dynamics can be described by
\begin{equation}\label{eq:lif-voltage}
\tau_{rc} \frac{dV(t)}{dt} = -(V(t)-V_r) + I(t)\,,
\end{equation}
where $I(t)$  is the input current, $V_{\text{r}}$ is a resting potential, and $\tau_{rc}$ is the time constant. Note that, if there is no input current, \eqref{eq:lif-voltage} describes how the membrane potential $V(t)$ decays towards its resting value $V_{\text{r}}$ with a characteristic time constant $\tau_{rc}$ \cite{gerstner2002spiking}. $I(t)$ implies the total stimulation that the neuron received, which can be included as an additional external input or stimulation from another connected neuron group. When postsynaptic currents are applied to a neuron, the membrane potential increases over time and, upon reaching a fixed threshold $\vartheta$, the neuron emits a spike, and the membrane potential resets to its resting value.


\subsection{Spike Encoding and Decoding}\label{sub:encoding}


To implement Belief Propagation in a spiking neural network, continuous probabilistic quantities must be represented by neural activity. In this work, these quantities correspond to Gaussian belief parameters, such as means and variances, that are propagated during message passing. Since spiking neurons communicate through discrete spikes rather than continuous-valued variables, a mechanism is required to encode real-valued means and variances into neural spike trains and decode them into mean and variance estimates.  

In computational neuroscience, spike coding schemes are commonly divided into rate coding and temporal coding. Rate coding represents information through the average firing rate of neurons, whereas temporal coding conveys information through the precise timing of spikes~\cite{gerstner2002spiking}.

In this work, we employ population coding, a form of rate coding in which information is represented by the joint activity of a neural population. This approach captures important characteristics of biological neural coding while remaining mathematically tractable for probabilistic modeling and analysis~\cite{wu2002population, zemel1998probabilistic, ma2006bayesian, sanger1996probability}. Population coding is also supported by biological
evidence, including place-cell representations in the hippocampus and neural activity patterns observed in cortical areas of primates and humans~\cite{o1971hippocampus, ceccarelli2023static, maunsell1983functional, shih2023electrophysiological}.

In population coding, different neurons become selective to different regions or directions of the represented space, and the represented variable is encoded by the collective activity of many neurons. Such distributed representations are robust to noise and enable accurate representation of continuous-valued variables.

Consider a neural population \(P\) consisting of \(N\) neurons that represents a
vector-valued variable \(x \in \mathbb{R}^d\). In our setting, \(x\) may denote a
belief parameter, for example, the mean of a Gaussian message.
Each neuron is assigned an encoder vector \(e_i \in \mathbb{R}^d\), which defines the preferred direction of that neuron in the represented space. The preferred direction corresponds to the orientation to which the neuron is most responsive, producing its highest firing activity when the represented variable is aligned with that direction.

We use the encoding and decoding method of the Neural Engineering Framework (NEF) \cite{eliasmith2003neural}. 
In this representation, variables (such as mean and variance) are encoded in the neurons' input currents, which then drive the neuron dynamics given by \eqref{eq:lif-voltage}. This current-based representation is required because the LIF neuron model operates on input currents rather than directly on abstract variables. The input current to neuron
\(i\) is given by
\begin{equation}
    I_i(x) = \alpha_i e_i^\top x + I_i^{\mathrm{bias}},
    \label{eq: current Jx from stim x}
\end{equation}
where \(\alpha_i\) is the gain and \(I_i^{\mathrm{bias}}\) is a constant bias
current. The term \(e_i^\top x\) measures the alignment between the represented
variable and the neuron's preferred direction. The gain \(\alpha_i\) scales this
projection and controls the sensitivity of the neuron, while
\(I_i^{\mathrm{bias}}\) shifts the baseline current and therefore determines the
neuron's firing threshold. The encoder vectors are randomly sampled and
normalized such that
\begin{equation}
    \|e_i\|_2 = 1 .
\end{equation}
This creates a diverse set of neuronal tuning curves, allowing the population to
represent continuous-valued variables in a distributed manner. 

After encoding, the value of the variable $x$ is represented by a distribution of spike activity across the neuronal population. To recover the represented value for downstream computation, neural activity is  decoded by 
\begin{equation}
    \hat{x} = D_P^\top R_P \,,
    \label{eq:nef_decoding}
\end{equation}
where  \(R_P \in \mathbb{R}^N\) denotes the population activity and
\(D_P \in \mathbb{R}^{N \times d}\) is the decoder matrix for population
\(P\). The decoder matrix is obtained by minimizing the reconstruction error over a set of sampled input points:

\begin{equation}
D_P = \arg\min_D \sum_x \left\|x - D^\top R_P(x)\right\|^2.
\label{eq: decoder}
\end{equation}

This least-squares optimization ensures that the population activity accurately approximates the represented variable over the desired domain.
The distributed nature of the decoder computation allows accurate representation despite nonlinear neuronal tuning and redundancy within the neural population~\cite{eliasmith2003neural}.

\subsection{Networks of Spiking Neurons}\label{sub:nef-networks}


We implement transformations using networks of LIF neurons by computing functions
through optimized synaptic connection weights. To achieve this, we follow the
principles of the NEF~\cite{eliasmith2003neural}.

Consider two neural populations, $P$ and $Q$, with $N_P$ and $N_Q$ neurons,
respectively. Population $P$ represents a variable $x \in \mathbb{R}^d$, and the
goal is to have population $Q$ represent a transformed variable
$z = f(x) \in \mathbb{R}^m$, where $f \colon \mathbb{R}^d \to \mathbb{R}^m$ is a
function.

To implement the transformation, we compute function-specific decoding vectors
$D_P^f \in \mathbb{R}^{N_P \times m}$ for population $P$ by solving the
least-squares problem
\begin{equation}
    D_P^f = \arg\min_{D} \sum_{x} \left\| f(x) - D^\top R_P(x) \right\|^2,
    \label{eq:function-decoder}
\end{equation}
where $R_P(x) \in \mathbb{R}^{N_P}$ is the activity vector of population $P$ in
response to input $x$. The columns $d_i^f \in \mathbb{R}^m$ of $D_P^f$ are the
per-neuron decoding vectors, such that $f(x) \approx (D_P^f)^\top R_P(x)$.
These decoding vectors are not applied as a separate computational stage in the
network; they are used solely to determine the synaptic weights between $P$ and
$Q$, as described next.

The synaptic weight from neuron $i$ in population $P$ to neuron $j$ in population
$Q$ is given by
\begin{equation}
    \omega_{ij} = \alpha_j \, e_j^\top d_i^f,
    \label{eq:synaptic-weight}
\end{equation}
where $\alpha_j$ is the gain of neuron $j$ in population $Q$ and
$e_j \in \mathbb{R}^m$ is its encoder vector. Each weight $\omega_{ij}$ therefore
folds the decoding from population $P$ and the encoding into population $Q$ into
a single scalar; there is no separate decoder or encoder block between the two
populations.

\begin{figure}[ht]
    \centering
    \includegraphics[width=0.85\linewidth]{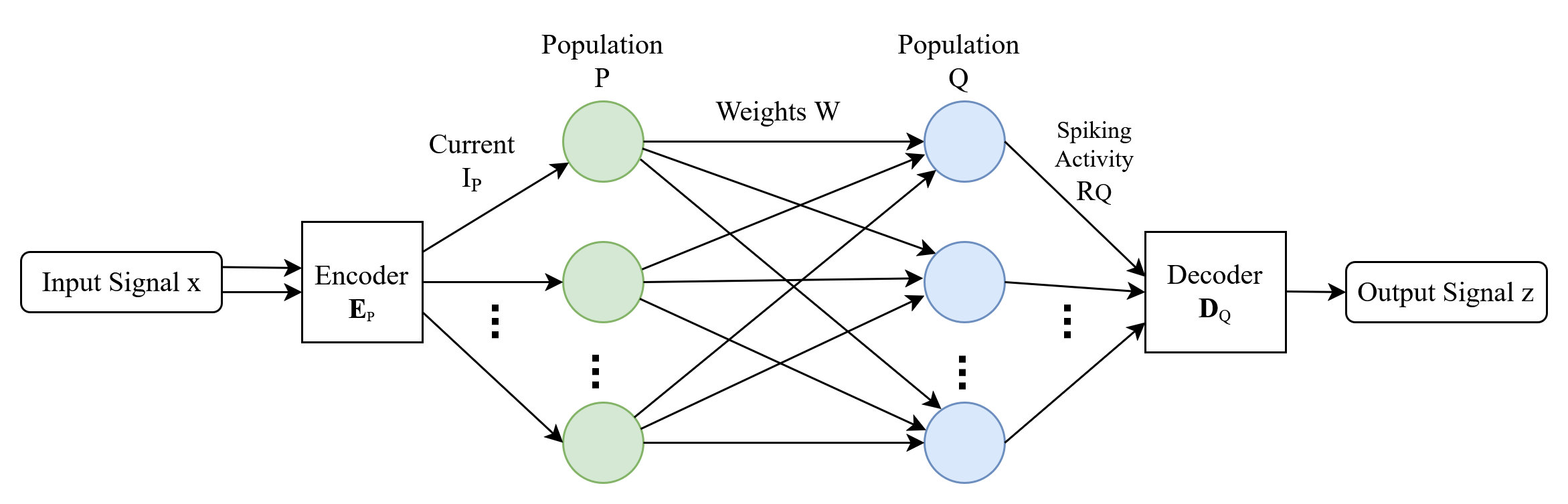}
    \caption{NEF-based function transformation between two spiking neural
    populations. Input signal $x$ is encoded into input currents $I_P$ for
    ensemble $P$. Synaptic weights $W$ connect population $P$ to population $Q$,
    implementing the transformation $f(x)$ implicitly. The decoder $D_Q$
    reconstructs the output signal $z \approx f(x)$ from the spiking
    activity $R_Q$.}
    \label{fig:nef-network}
\end{figure}

In practice, each neuron in population $P$ emits a spike train, and the
postsynaptic input to neuron $j$ in population $Q$ is formed by filtering these
spike trains through a synaptic impulse response $k(t)$. Let
$a_i(t) = \sum_n k(t - t_n^{(i)})$ denote the filtered activity of neuron
$i$, where $\{t_n^{(i)}\}$ are its spike times and $k(t)$ is an exponential
kernel with synaptic time constant $\tau_s$. The resulting input current to
neuron $j$ in population $Q$ is
\begin{equation}
    I_j(t) = \sum_{i=1}^{N_P} \omega_{ij} \, a_i(t) + I_j^{\mathrm{bias}}
            = \alpha_j \, e_j^\top \!\left( \sum_{i=1}^{N_P} d_i^f \,
              \hat{a}_i(t) \right) + I_j^{\mathrm{bias}}.
    \label{eq:postsynaptic-current}
\end{equation}
Since $\sum_i d_i^f \hat{a}_i(t) \approx f(x(t))$, the bracketed term
approximates the decoded function value, and \eqref{eq:postsynaptic-current}
reduces to
\begin{equation}
    I_j(t) \approx \alpha_j \, e_j^\top f(x(t)) + I_j^{\mathrm{bias}},
    \label{eq:nef-transform-current}
\end{equation}
which has exactly the form of \eqref{eq: current Jx from stim x} with $x$
replaced by $f(x)$. Figure~\ref{fig:nef-network} illustrates this two-population
architecture. Consequently, population $Q$ encodes the transformed variable
$f(x)$ according to the same NEF scheme described in Section~\ref{sub:encoding}.

This construction allows arbitrary smooth functions to be approximated between
neural populations purely through the synaptic weight matrix $W$, with no explicit intermediate computation. The decoders $D_P^f$ are computed offline
during network construction and combined with the encoders of population $Q$ to
set the weights; once the network is running, only spikes and synaptic currents
are exchanged~\cite{eliasmith2003neural, bekolay2014nengo}.

\section{Methods}\label{sec:model}

In this section, we present the model specification, NEF realization, and inference method for spike-based belief propagation on a factor graph for the Mountain Car problem.  

\subsection{A State Space Model for Mountain Car Dynamics}\label{sec:gen-model-mountain-car-agent}

We present a realization of the extended generative model \eqref{eq:complete-model} for the Mountain Car agent.

First, we describe the state transition model $p(s_t|s_{t-1},u_t)$.  The Mountain Car's state is $s_t = [x_t,\; \dot{x}_t]^\top$, where $x_t$ is the car's position and $\dot{x}_t$ its velocity. Following~\cite{ueltzhoffer2018deep}, the discrete-time transition dynamics are given by
\begin{subequations}\label{eq:mc-position}
    \begin{align} 
\dot{x}_t &= \dot{x}_{t-1} + F_g(x_{t-1}) + F_f(\dot{x}_{t-1}) + F_a(u_t) 
\label{eq:mc-velocity} \\ 
x_t &= x_{t-1} + \dot{x}_t, 
\end{align} 
\end{subequations}

where $F_g$, $F_f$, and $F_a$ are the gravitational, friction, and engine force components. The engine force is a bounded function of the control action 
\begin{align}\label{eq:F_a}
F_a(u_t) = F_{\lim}\,\tanh(u_t)\,.
\end{align}
The friction force is linear in velocity, 
\begin{align} 
F_f(\dot{x}) = -0.1\,\dot{x} \,,
\end{align} 
and 
\begin{equation}\label{eq:F_g}
F_g(x)=
\begin{cases}
-0.05\,(2x+1), & \text{if } x < 0,\\[6pt]
-0.05\left[(1+5x^2)^{-1/2}
+ x^2(1+5x^2)^{-3/2}
+ \frac{1}{16}x^4\right], & \text{otherwise}.
\end{cases}
\end{equation}
determines the horizontal gravitational force of the hilly landscape~\cite{van2019simulating}.


The full state transition model is given by
\begin{align}
p(s_k | s_{k-1}, u_k) =
\mathcal{N}\left( s_k \,|\,
\tilde{g}(s_{k-1},u_k),
\,10^{-4}I
\right)\,,
\end{align}
where $\tilde{g}$ is the spike-based realization of 
\begin{equation}\label{eq:g-mountain-car}
g(s_{t-1}, u_t) =
\begin{bmatrix}
x_{t-1} + \dot{x}_{t-1} + F_g(x_{t-1}) + F_f(\dot{x}_{t-1}) + F_a(u_t) \\
\dot{x}_{t-1} + F_g(x_{t-1}) + F_f(\dot{x}_{t-1}) + F_a(u_t)
\end{bmatrix}\,.
\end{equation}
Equation~\ref{eq:g-mountain-car} follows from \eqref{eq:mc-position}--\eqref{eq:F_g}.

We assume we observe noisy versions of the states. The observation model of the Mountain Car agent is then given by 

\begin{equation}
p(y_k |s_k) = \mathcal{N}\left( y_k \,|\, s_k, \,10^{-4}I \right) \,.
\end{equation}

We also need to specify the preference prior for future observations. We define the prior $p'(y_k)$ for preferred future observations as

\begin{equation}\label{eq:target-observations} 
p'(y_k)= \begin{cases} 
\mathcal{N}(y_k \mid y^*, 10^{8} I), & \text{if } k < T -t,\\ 
\mathcal{N}(y_k \mid y^*, 10^{-8} I), & T-t \leq k \leq T. 
\end{cases} 
\end{equation}
Here, $y^*$ denotes the preferred future position state, that is, the target parking location. For a given time step $t$, the index $k \in \{1,\ldots,T\}$ denotes a future step within the planning horizon. The preference prior is assigned a very large covariance for $k<T-t$, resulting in a broad distribution that permits exploration. The remaining observations are assigned a very small covariance, producing a sharp distribution centered at $y^*$. As time progresses, the number of the broad prior decreases, while an increasing portion of the horizon becomes strongly constrained toward the target state. Consequently, the agent is allowed to explore early in the trajectory but is progressively driven toward the target observation as the end of the planning horizon approaches. For more details on this approach, we refer the reader to~\cite{van2019simulating}.

Finally, to simulate inference in \eqref{eq:complete-model}, we need to specify the priors
\begin{subequations}
 \begin{align}
  p(s_0) &= \mathcal{N}\left( s_0 \,|\, (-0.5,\,0), \,10^{-12}I \right) \\
  p(u_k) &= \mathcal{N}\left( u_k \,|\, 0, \,10^{12} \right)\,.
\end{align}   
\end{subequations}

\subsection{NEF Realization of Mountain Car Dynamics}\label{subsection:SNN_node}
To implement the Mountain Car dynamics in the NEF framework, we use the construction described in Section~\ref{sub:nef-networks}. Separate neural populations are used to approximate $F_g$, $F_f$, and $F_a$, with respectively $N_g$, $N_f$, and $N_a$ neurons. Because $F_g$ is approximately linear for $x < 0$ and nonlinear for $x \geq 0$, it is implemented using two separate populations of size $N_g$, whose outputs are selected according to the sign of $x$.

For each population, the encoder matrix, gain parameters $\alpha_i$, bias currents $I_i^{\mathrm{bias}}$, and decoder matrix are defined according to the NEF formulation. In the current work, these parameters are generated automatically using the \texttt{Nengo} toolbox~\cite{bekolay2014nengo}, which implements the Neural Engineering Framework and initializes heterogeneous neuron properties over the represented input space. Specifically, \texttt{Nengo} samples neuron intercepts and maximum firing rates and computes the corresponding gain and bias parameters according to the NEF formulation. As a result, gain values vary across neurons, yielding heterogeneous tuning curves within the population.

The encoder determines the preferred input direction of each neuron, while $\alpha$ and $I_{\mathrm{bias}}$ control the neuron firing threshold and response strength. This diversity enables the neural populations to accurately represent the nonlinear functions underlying the transition dynamics.

We also used \texttt{Nengo} to automate the optimization tasks in \eqref{eq: decoder} for computing the decoder matrices and connection weights for each network. All computations are performed offline, and the resulting fixed weights and matrices are later used in the spike-based node function, which is integrated into the complete factor graph framework, as described in the following section.

\subsection{Nonlinear Message Approximation}\label{subsection:UKF}

\begin{figure}[t]
    \centering
    \includegraphics[width=\linewidth]{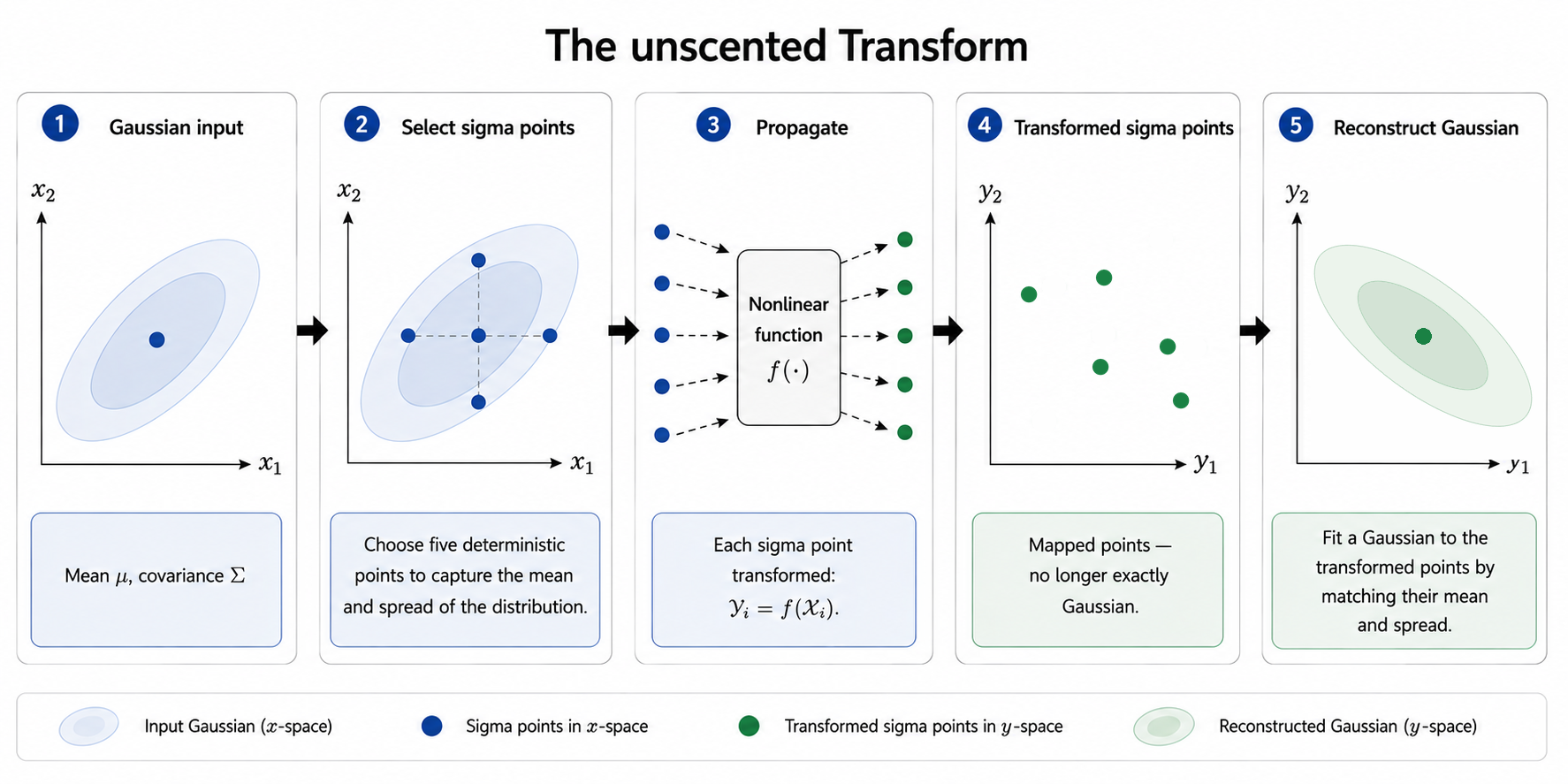}
    \caption{Illustration of the unscented transform. A Gaussian distribution is represented by a deterministic set of sigma points, which are propagated through a nonlinear function. The transformed points are then used to reconstruct a Gaussian approximation of the resulting distribution.}
    \label{fig:ukf-transform}
\end{figure}

Nonlinear factor nodes generally transform Gaussian messages into non-Gaussian distributions. Since our BP framework represents messages as Gaussians, these distributions must be approximated to maintain tractable inference. To this end, we use the Unscented Kalman Filter \cite{julier1997new} as a local message-approximation method. As illustrated in Fig.~\ref{fig:ukf-transform}, this transformation first represents a Gaussian distribution using a deterministic set of sigma points. These points are propagated through the nonlinear transformation, and the transformed points are then used to reconstruct a Gaussian approximation of the resulting distribution. This allows nonlinear message updates to be incorporated into the Gaussian BP framework without explicit linearization. In our implementation, these UKF-based message updates are automated using the \texttt{RxInfer} framework~\cite{bagaev2023reactive}.

\subsection{Inference Procedure}\label{subsection:model-inference}

The overall inference and control procedure is summarized in Algorithm~\ref{alg:inference}. We follow the message passing procedure for active inference agents as detailed in \cite{van2019simulating}. At each time step, the agent executes the current control action and receives a new observation from the environment. The observation is processed by the factor graph through belief propagation, yielding an updated posterior belief over the latent states. The inferred states are then used to update the posterior distribution over future control actions by performing message passing on the extended planning model. The first action of the resulting control sequence is executed, and the procedure is repeated at the next time step.

\begin{algorithm}[bt] 
\caption{Inference in the Mountain Car Agent}\label{alg:inference}
\begin{algorithmic}
\Require generative model from section~\ref{sec:gen-model-mountain-car-agent}

\State $t \gets 0$
\While{$ t \leq T$}

\State execute action $u_t \sim p(u_t|y_{1:t-1})$
\State observe $y_t$
\State  infer state update $p(s_t|y_{1:t})$ \quad (message passing in model, Fig.~\ref{fig:FFG-state_update})
\State infer control update $p(u_{t+1:T}|y_{1:t})$ \quad (message passing in ext. model, Fig.~\ref{fig:FFG-KF})
\State $t \gets t + 1$

\EndWhile

\end{algorithmic}
\end{algorithm}

\section{Experimental Validation}\label{sec:results}

We evaluated whether the proposed spike-based realization of the nonlinear state-transition factor can be integrated into a belief propagation framework without degrading the control behavior of the original active inference agent. As a reference, we compare our implementation against the active inference controller of Van de Laar and de Vries \cite{van2019simulating} (the "reference" implementation), which performs inference using conventional numerical computations rather than spike-based neural dynamics.

\begin{figure}[tb!]
    \centering

    \begin{subfigure}[t]{0.48\textwidth}
        \centering
        \includegraphics[width=\linewidth]{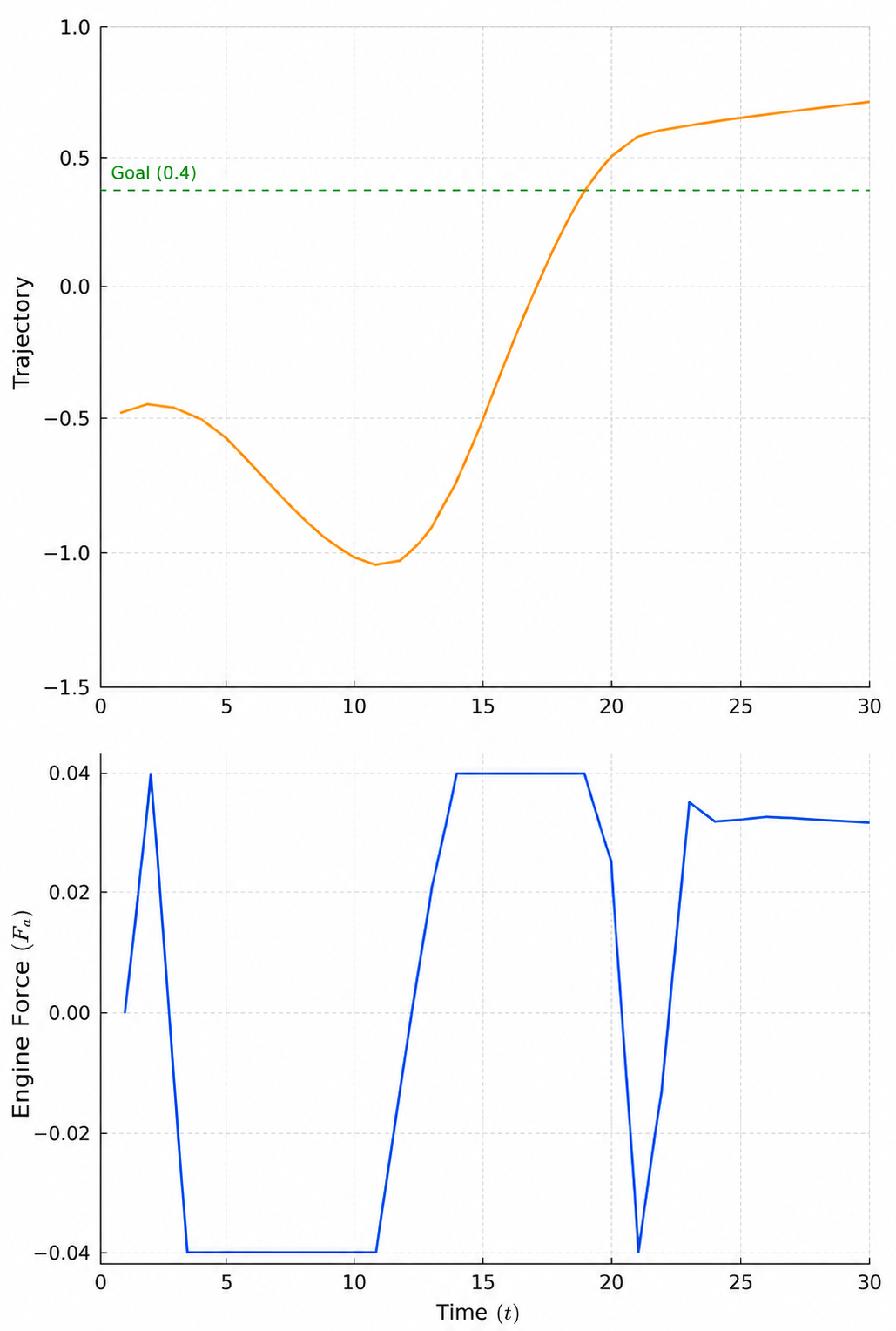}
        \caption{Reference Active inference controller \cite{van2019simulating}.}
        \label{fig:active-inference}
    \end{subfigure}
    \hfill
    \begin{subfigure}[t]{0.48\textwidth}
        \centering
        \includegraphics[width=\linewidth]{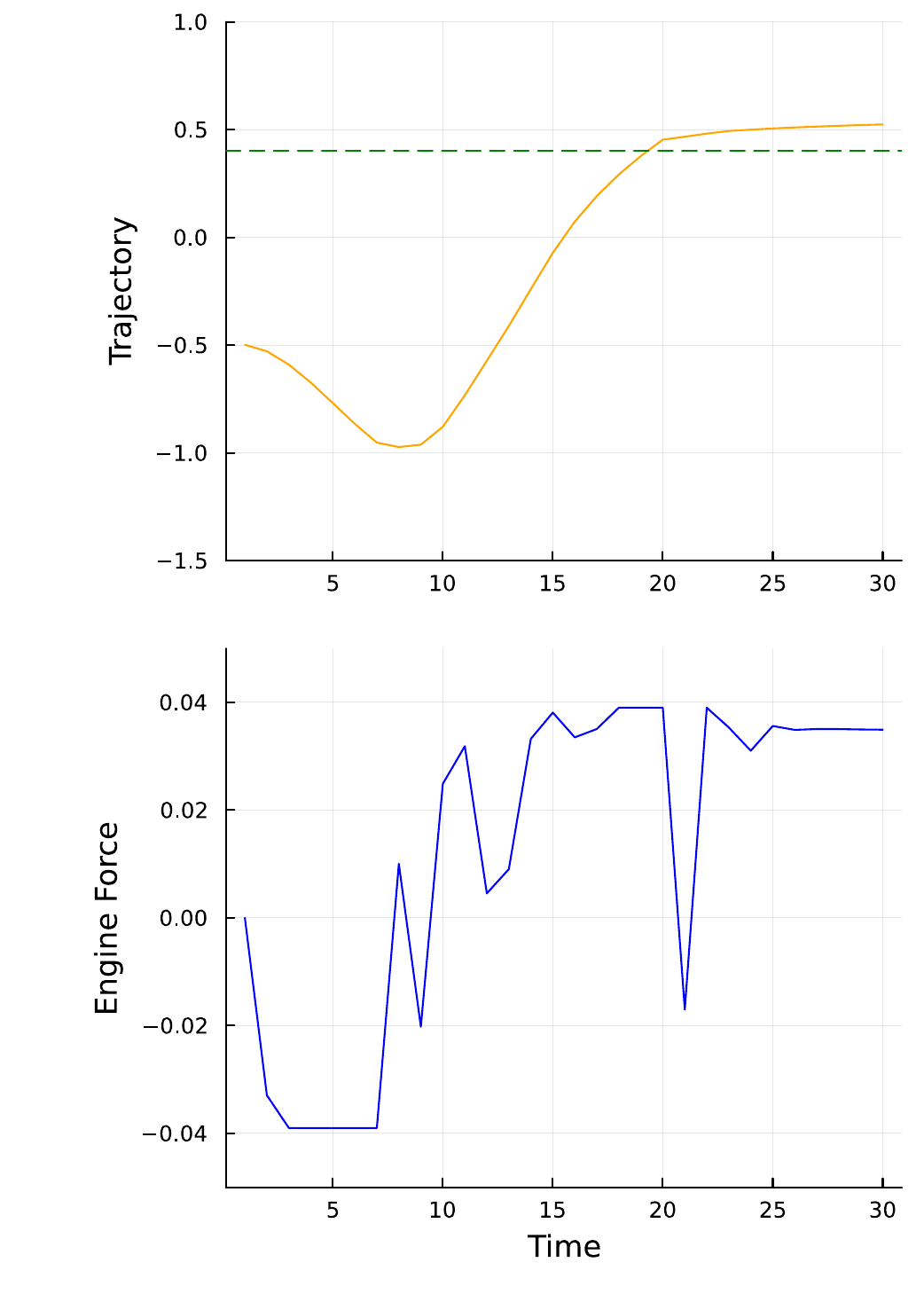}
        \caption{Proposed (spiking-based) method.}
        \label{fig:proposed-method}
    \end{subfigure}

    \caption{Comparison of the reference active inference controller (left), versus the proposed controller (right) for the Mountain Car environment. The top panel shows the vehicle position over time, with the dashed horizontal line indicating the goal position. The bottom panel shows the corresponding engine force applied at each time step. Both methods exhibit the characteristic Mountain Car strategy of initially moving away from the goal to build momentum before ascending the hill. In both cases, the vehicle successfully reaches and overshoots the goal position, demonstrating successful task completion. }
    \label{fig:comparison}
\end{figure}

We used \texttt{Nengo}~\cite{bekolay2014nengo} to implement the spike-based realization of the nonlinear functions and to automate the encoding and decoding procedures between continuous-valued variables and neural population activity. \texttt{Nengo} was also used to compute the decoder matrices and synaptic connection weights according to \eqref{eq: decoder} and \eqref{eq:synaptic-weight}. The message-passing procedure and UKF-based message approximations were implemented using the \texttt{RxInfer} framework~\cite{bagaev2023reactive}, which provides an event-driven realization of belief propagation.

The parameters used in the experiment were $u_0 = 0$, $y^* = 0.4$, $N_g = 300$, $N_f = 100$, $N_a = 100$, and $F_{\mathrm{lim}} = 0.04$.

Figure~\ref{fig:comparison} compares the behavior of the reference active inference controller and the proposed spike-based implementation. The upper plots show the vehicle's position as a function of time, while the lower plots show the corresponding engine control actions. The purpose of this comparison is not to exactly reproduce the control trajectory of the reference implementation, but to verify that the spike-based realization preserves the qualitative control behavior. As shown in Fig.~\ref{fig:comparison}, both controllers initially move away from the goal to accumulate momentum and subsequently drive the vehicle toward the target position. In both cases, the vehicle reaches and overshoots the goal position, demonstrating that the proposed spike-based implementation successfully reproduces the intended active inference control strategy.

\section{Conclusions and Future Work}\label{sec:conclusions}


This paper presented a spike-based message-passing framework for Bayesian control based on factor graphs and spiking neural networks. By combining belief propagation with SNNs, we developed a biology-inspired implementation of a nonlinear, multivariate state transition factor for a Mountain Car agent. The proposed approach integrates spike-based computation with probabilistic inference, enabling state estimation and control through local message passing on a factor graph.

The resulting model was implemented using \texttt{Nengo} and integrated into an event-driven inference framework using \texttt{RxInfer}. Experimental results demonstrated that the proposed controller successfully solves the Mountain Car problem. These results indicate that spike-based neural dynamics can support nonlinear Bayesian inference and control in a biologically inspired manner.

Interesting directions for future work include extending the framework to a fully spike-based implementation, in which both the factor nodes and the message-passing algorithm are realized through spiking neural dynamics. Such an implementation would enable deployment on neuromorphic hardware and facilitate the evaluation of its energy efficiency.

Another promising direction is to replace the hand-crafted state-transition model with a learned dynamics model. Rather than relying on predefined equations, the state-transition factor could be learned directly from interactions with the environment, allowing the framework to be applied to more complex and less structured control problems.

\begin{credits}
\subsubsection{\ackname}
We gratefully acknowledge funding for this study by the Eindhoven Artificial Intelligence Systems Institute (EAISI) at Eindhoven University of Technology.
\end{credits}

\bibliographystyle{unsrt}

\bibliography{content/references}

\end{document}